\documentclass[preprint,5p]{elsarticle}

\usepackage{prletters}
\usepackage{amsmath,amssymb}
\usepackage{hyperref}
\hypersetup{
    colorlinks=true,
    breaklinks=true,
    linkcolor=red,
    filecolor=magenta,      
    urlcolor=blue,
    citecolor=blue
}

\begin{document}

\begin{frontmatter}


\title{Confidence Estimation for Object Detection in Document Images}

\author[1,2]{Mélodie Boillet\corref{cor1}}
\ead{boillet@teklia.com}
\author[1,2]{Christopher Kermorvant}
\ead{kermorvant@teklia.com}
\author[2]{Thierry Paquet}
\ead{thierry.paquet@univ-rouen.com}

\cortext[cor1]{Corresponding author}
\address[1]{TEKLIA, Paris, France}
\address[2]{LITIS, Normandie University, Rouen, France}


\begin{abstract}

Deep neural networks are becoming increasingly powerful and large and always require more labelled data to be trained. However, since annotating data is time-consuming, it is now necessary to develop systems that show good performance while learning on a limited amount of data. These data must be correctly chosen to obtain models that are still efficient. For this, the systems must be able to determine which data should be annotated to achieve the best results.

In this paper, we propose four estimators to estimate the confidence of object detection predictions. The first two are based on Monte Carlo dropout, the third one on descriptive statistics and the last one on the detector posterior probabilities. In the active learning framework, the three first estimators show a significant improvement in performance for the detection of document physical pages and text lines compared to a random selection of images. We also show that the proposed estimator based on descriptive statistics can replace MC dropout, reducing the computational cost without compromising the performances. 


\end{abstract}


\begin{keyword}
Confidence Estimation \sep Document Object Detection \sep Active Learning
\end{keyword}

\end{frontmatter}


\section{Introduction}

Despite the remarkable performance of deep neural networks in many application domains, their successful introduction in real-world production systems requires that they not only perform well but also have some capacities to assess the certainty or uncertainty of their decisions. This is particularly important for medical image or autonomous driving related applications. But the problem also arises in the case of domain adaptation of deep neural networks, where one expects to provide the system with as few new labelled examples as possible to adapt the system to the new domain. Choosing the relevant examples for human labelling is crucial to allow a successful adaptation. This framework known as active learning requires that a first system can perform the task while automatically assessing its confidence on new unseen data, so that the less confident decisions can be submitted to a human operator for manual labelling, whereas the more confident decisions made by the system would be kept as is to provide an automatic labelling. In this paper, we aim at developing confidence measures for document object detection model adaptation within an active learning framework, so as to reduce the human annotation effort to a minimum. \\

Object detection neural networks output probabilities that could be directly used as confidence estimates. However, it has been shown that these probabilities are often overconfident estimators that give high confidence even on erroneous predictions \cite{nguyen2015}. To address this problem, several studies have been conducted to design better estimators.

Within the active learning framework, one of the first proposed approaches to select the samples to be manually annotated was based on linear Support Vector Machines (SVM) \cite{tong2002}. Another popular approach is uncertainty sampling \cite{settles2008} where the samples leading to predictions with high uncertainty are selected. To quantify the uncertainty, several measures based on posterior probabilities have been proposed such as entropy or least confidence score \cite{brust2019}.

To model the uncertainty of neural networks decisions, some other approaches have been proposed such as the Monte Carlo (MC) dropout \cite{mcdropout}. Instead of computing a single prediction at test time, the network is asked to provide multiple predictions with dropout, the distribution of which is then analyzed to derive a confidence estimate of the non dropout prediction. This technique, which approaches Bayesian deep learning models, has been used for many tasks. It has often proved to be efficient for classification to choose the data to be labelled \cite{gal2017}. In \cite{bayesianUNet}, MC dropout is used to estimate the uncertainty in semantic image segmentation. 
In addition, MC dropout has been used as a class probabilities regularization technique to get an improved ordinal ranking of the predictions \cite{moon2020crl}.

Some other works make use of deep confidence estimation models independent of the detection model. In \cite{granell2021}, an adversarial network is trained at the same time as the baseline detection model. The adversarial network is trained to estimate how close the predictions are to the ground-truth. It is worth noting that most of the aforementioned works focus on classification tasks. Despite the large number of works focusing on the implementation of new object detection systems, there is little to no literature focusing on the confidence estimation for this task. \\

The aim of this work is to build a confidence estimator for object detection in document images within an active learning scenario. For this purpose, we investigate four confidence estimators. The first one consists in using the class posterior probabilities of the detection model to estimate the confidence. In the second approach, we propose two confidence estimators inspired by the Monte Carlo dropout that consists in building confidence estimates using dropout at test time.
The main advantage of this approach is that no additional training is required as long as the model has been trained with dropout layers. It can be applied to already trained models without any modification. This approach, however, is computationally expensive, and therefore our last proposal consists in building a dedicated system that can predict a confidence estimate with only one forward step during inference. Being independent of the predictor, this system requires a specific training phase. \\

Our contributions are as follows:
\begin{itemize}
    \item We propose two new confidence estimators based on Monte Carlo dropout for object detection tasks.
    \item We propose a new confidence estimator, specifically designed for object detection, which is  based on object descriptive statistics that estimates the mean Average Precision (mAP) of the predictions.
    \item We present an estimator based on class posterior probabilities.
    \item On two document analysis tasks (page detection and text line detection), we compare the estimators through extensive experiments and notably in an active learning setting where a good confidence estimate is crucial. We show that with half as much data, they allow achieving much better results than those obtained with a random selection of the adaptation samples.
\end{itemize}

\fussy
This paper is organized as follows: Section \ref{sec:conf_est_systems} presents the proposed confidence estimators. Section \ref{sec:experimental_setup} presents the setup used for the experiments: data, training details of the detection models and those of the confidence estimators. Finally, in Section \ref{sec:experimental_results} we present and discuss the results obtained.
\sloppy

\section{Proposed confidence estimators}
\label{sec:conf_est_systems}

In the following, we present the four confidence estimators we propose: the first one is a class posterior probabilities-based estimator; the following two are based on Monte Carlo dropout \cite{mcdropout} and the last one on descriptive statistics.

\subsection{Estimator based on posterior probabilities}

Since object detection neural networks output probabilities, a straightforward confidence estimate is the Posterior probabilities-based Confidence Estimator (denoted in the following as PCE). For each predicted object, the pixel probabilities output by the detection model are first averaged. Then the PCE score is computed and corresponds to the average of all the objects' probabilities.

\subsection{Estimators based on Monte Carlo dropout}

Estimating the confidence of a prediction with MC dropout consists in computing \textit{N} forward steps on the same observation and analyzing the distribution of the predictions. The variance between the \textit{N} predictions is an indicator of the network uncertainty and can thus be considered as a confidence estimate. In this paper, we propose two scores summarizing the variance of the predictions: The Dropout Average Precision (DAP) and the Dropout Object Variance (DOV). \\

The mean Average Precision (mAP)\footnote{\url{https://gitlab.com/teklia/dla/document_image_segmentation_scoring}} used in the PASCAL VOC Challenges \cite{everingham2010} allows evaluating a prediction at object level compared to a ground-truth detection result. The advantage of this metric is that it considers the size and the position of the predicted objects, since it relies on an Intersection-over-Union object matching. Inspired by this metric, we derive the Dropout Average Precision (DAP) estimator which is computed by considering every prediction pair (($p_i$, $p_j$) where $p_i$ and $p_j$ are two distinct predictions of the same image with $i, j \in N$ and $i \neq j$) and computing the mAP for each pair, one prediction being considered as ground-truth arbitrarily. The DAP is the average of all the mAP scores (see Equation \ref{eq:dap}) where a high DAP indicates that the \textit{N} predictions are very similar and is likely an indicator of correct detection.

\begin{flalign}
    DAP & = \frac{1}{N^2-N} \times \sum_{i=1, j=1, i \neq j}^{N}{mAP(p_i, p_j)}
    \label{eq:dap}
\end{flalign}

The second estimator we propose is based only on the variance of the number of predicted objects among the dropout predictions, thus the name Dropout Object Variance (DOV) of this estimator. Indeed, when the model is not very confident, a highly variable number of objects is predicted with many small objects around the main one (as shown on the right image of Figure \ref{fig:variance_images}). To get a single value, we take the variance of the number of objects in the dropout predictions as shown in Equation \ref{eq:dov} where $n_i$ is the number of objects in the prediction $p_i$. A DOV of 0 indicates that all the predictions have the same number of objects, and is likely an indicator of correct detection.

\begin{flalign}
    DOV & = \frac{1}{N} \times \sum_{i=1}^{N}{(n_i - \overline{n})^{2}} \text{~~ with ~} \overline{n} = \frac{1}{N} \times \sum_{i=1}^{N}{n_i}
    \label{eq:dov}
\end{flalign}

These two scores are able to estimate the confidence of a prediction with a variable number of objects. Indeed, the number of objects to detect depends on the task and can be highly variable: one or two for page detection and several tens for text line detection.

\subsection{Estimator based on object statistics}

In this section, we adopt a more standard approach for confidence estimation based on regression. We design a specific system that analyzes a detection result and estimates the mAP, as no ground-truth is available at test time. Unlike our first proposal, the system being independent of the detector, this approach can operate with any type of detector.

\subsubsection{Object descriptive statistics}
\label{sec:features}

At the output of the detection model, we have class posterior probabilities for each pixel to belong to the object or background class. First, pixels are assigned to the class with the highest probability and then the connected components are detected, which leads to several predicted objects for a given image. Then we extract the bounding polygons of the connected components as well as their bounding rectangles. From this information, we compute the following eight object features:

\begin{itemize}
    \item Height and width ratios between each predicted object's bounding rectangle and the image height and width;
    \item Ratio between the height and width of the bounding rectangle of each predicted object;
    \item For each predicted object, ratios between the area of its bounding polygon and the area of the image, the area of its bounding polygon and the area of its bounding rectangle, the area of its bounding rectangle and the area of the image;
    \item Distances between the centroids of all bounding rectangles in height (normalized by the image height) and in width (normalized by image width). The distances are computed by considering each pair of bounding rectangles.
\end{itemize}

These features allow describing the sizes, shapes, and positions of the detected objects in document images. For a given image, each feature is computed for each detected object, whose resulting values are grouped into \textit{B} bins to provide a histogram for each feature. The histograms are then concatenated to form a single object statistic feature vector of size 8$\times$\textit{B}. These statistics are then used to train a regressor.

\subsubsection{mAP-RFR}

To build the confidence estimator, we chose to estimate the mAP of the predictions because it has been shown to be more meaningful than the IoU \cite{boillet2022}. To estimate the mAP of a prediction, several regression methods can be used, such as Support Vector Regression (SVR) or Random Forest Regressor (RFR). In our experiments, we used RFR as it showed the best results in our preliminary works. After applying the regressor, no further processing is needed, since it directly outputs a single score that is considered as the confidence estimate. In the following, this estimator is referred to as mAP-RFR.

\section{Experimental setup}
\label{sec:experimental_setup}

The estimators are tested and compared on two tasks: physical document page detection and handwritten text line detection. Physical page detection is a rather simple task, since only one or two objects must be detected. Handwritten text line detection is a more complex task since document pages may contain a variable and possibly large number of text lines, very different in shape and positions.

\subsection{Experimental data}
\label{sec:data}

For the experiments on page detection, we used the cBAD \cite{gruning_cbad} and Horae \cite{boillet2019} datasets. Our goal is to adapt the detection model pre-trained on cBAD to the Horae document images by annotating as little data as possible. For the text line detection task, our goal is to adapt a pre-trained and generic line detection model to a new unseen set of documents, namely the Hugin-Munin dataset \cite{hugin-munin}. \\

\textit{The cBAD dataset} \cite{gruning_cbad} contains 2,035 images of handwritten documents that have been used during the cBAD competitions. The dataset has been annotated at single and double page levels\footnote{\url{https://github.com/ctensmeyer/pagenet}} \cite{pagenet2017}. For the following experiments, we predict at the single page level, leading to two objects for the images showing a double page document.
In addition, the \textit{abnormal} images have been removed since their annotations were not accurate enough. In the following, this version of the dataset is denoted as cBAD* and consists of 1,630 training, 200 validation and 199 test images with respectively 1,801, 221 and 219 single pages.

\textit{The Horae dataset} \cite{boillet2019} consists of 572 annotated images from 500 medieval books of hours. The full Horae corpus is composed of 1,158 books of hours presenting a high diversity of non-annotated document images in digitization types, backgrounds, and shapes. This corpus is used to compare the different estimators when used in a real active learning framework. In addition, since the original test set contains only 30 images, to obtain more robust results in the following experiments, we extended this test set by annotating 300 additional images randomly selected from the 1,158 books, representing 364 single pages. This test set is denoted Horae-test-300.

\textit{The Hugin-Munin dataset} \cite{hugin-munin} consists of annotated pages from private correspondences and diaries of 12 Norwegian artists written between 1820 and 1950. The documents have been annotated at line level with their corresponding transcriptions. This dataset contains 691 training, 85 validation and 73 test images. 

\subsection{Object detection training}
\label{sec:page_segmentation_training}

For our experiments, we used the Doc-UFCN\footnote{\url{https://pypi.org/project/doc-ufcn/}} \cite{boillet2020} system as object detector because it showed good performance for object detection on historical documents while having a reduced inference time compared to other systems. This system predicts pixel-level probabilities, and the pixels are assigned to the class with the highest probability. They are then grouped into connected components to form objects and those with an area smaller than a threshold \textit{t}=50 pixels are removed. Several values for this threshold have been tested in previous works \cite{boillet2022}, this one giving the best performance. \\

For both tasks, the pre-trained models (denoted as baseline in the following) are trained with the images resized such that their largest side is 768px, keeping their aspect ratio. A pre-processing is applied to the training labels to prevent the annotated areas from touching when resizing the images \cite{boillet2022}. The models are trained for 150 epochs with a learning rate of 5e-3 and the Adam optimizer. The weights that led to the lowest validation loss during training are kept as the best.

For the page detection task, the baseline model is trained on cBAD* images. It shows an IoU of 97\% and a mAP of 94\% on cBAD*. However, there is still room for improvement on the Horae-test-300 images, since the mAP is about 60\%. In the following, the images with the lowest estimated confidence scores in the Horae corpus are annotated to improve the detection evaluation on Horae-test-300.

For text line detection, we trained a generic text line segmentation model and then the confidence estimators on many datasets. In this regard, we have collected 19 mostly public databases including historical and modern documents \cite{boillet2022}. Altogether this dataset contains 9,432 training, 1,907 validation and 6,669 test images which corresponds to 374,316 training, 85,208 validation and 190,502 test annotated lines. 
This generic model applied to the Hugin-Munin test set was evaluated at 48\% IoU and 21\% mAP. This quite low result was expected since the documents are way more complex than those used during pre-training. \\

In addition to the standard segmentation metrics, text line models are evaluated using goal-directed metrics, namely the page-level CER and WER \cite{boillet2022}. For that purpose, a Handwritten Text Recognizer (HTR) based on Kaldi \cite{arora2019} was trained on the Hugin-Munin transcribed lines. We chose this HTR because it is an out-of-the-box package that generally performs reasonably well in most use cases and has shown competitive performances on Hugin-Munin documents  \cite{hugin-munin}.

The trained HTR model is applied to all the lines predicted by Doc-UFCN which are ordered by their centroid from the top-left corner of the page to the bottom-right corner. The predicted texts are concatenated in this same order to provide a single page-level transcription. Manual transcriptions are ordered using the same method and page level CER and WER are computed. The baseline detection model gives about 24\% CER on Hugin-Munin.
In addition, we compute the WordCountFMeasure (WCFM) \cite{bow2} which evaluates HTR models based on the number of correctly retrieved words, regardless of their position. We used the PRImA Text Evaluation Toolkit\footnote{\url{https://www.primaresearch.org/tools/PerformanceEvaluation}} to compute the WCFM scores. Kaldi gives a WCFM of 59\% compared to the manual transcriptions. These CER and WCFM values indicate that the lines detected by the baseline model are not the best input for the HTR model. They may reflect misplaced detected lines (no text), too thin lines (cut text) or missed lines.

\subsection{Confidence estimators training}
\label{sec:scoring_systems_training_and_evaluation}

No additional training is required for the estimators based on MC dropout, since only the object detection models presented in Section \ref{sec:page_segmentation_training} are used to estimate the confidence. In contrast, the regressors have to be trained. Those are trained on the same data as the detection models, whose statistics are presented in Sections \ref{sec:data} and \ref{sec:page_segmentation_training}. Thus, a first regressor is trained on the cBAD* data for page detection and a second is trained on the same 19 databases as the generic line detection model.

First the object detection model is applied to all the images (cBAD* for the page detection and the 19 datasets for the line detection) which allows calculating the statistics. Since the datasets are annotated, the detection model is then evaluated on each image separately, providing an IoU and mAP for each image. These mAP values are used as the target for training the regressors.

To train the regression models, we used the RandomForestRegressor of the scikit-learn toolkit using the default parameters. The regression models show low Mean Square Errors (MSE) on the training datasets (0.0164 MSE on the test set of cBAD*).

\section{Experimental results and discussion}
\label{sec:experimental_results}

In this section, we evaluate and compare the confidence estimators using reject curves and then compare their performance when integrated within an active learning framework.

\begin{figure}[t]
  \centering
    \includegraphics[width=\linewidth]{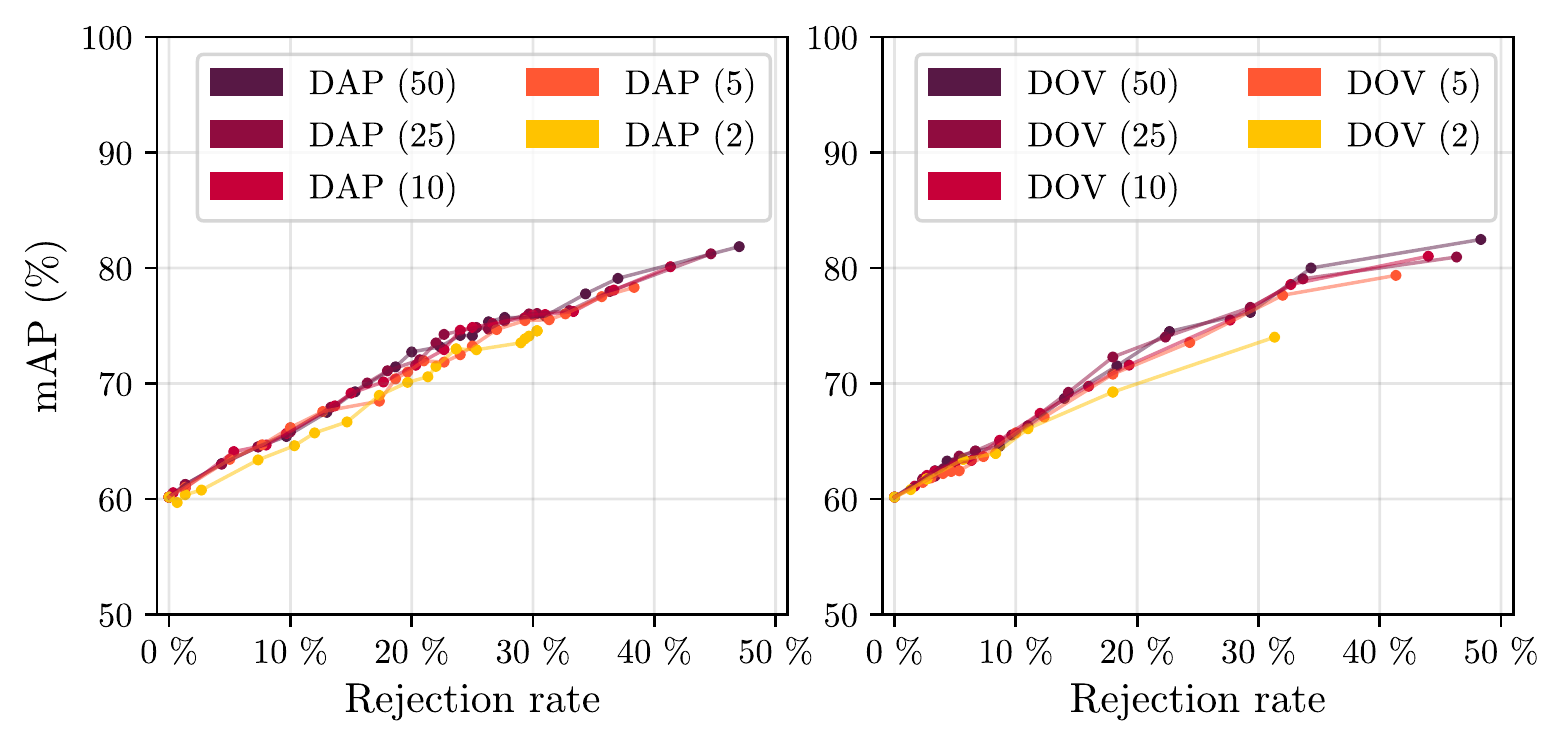}
    \caption{Reject curves showing the evolution of the baseline model performance on Horae-test-300 using DAP (left) and DOV (right) estimators regarding the number of dropout predictions \textit{N}.}
    \label{fig:rejection_curves_page_horae_N}
\end{figure}

\subsection{Number of dropout predictions}
Before further experimentation, we need to define the number of predictions \textit{N} to compute for the estimators based on MC dropout (DAP and DOV). Figure \ref{fig:rejection_curves_page_horae_N} shows the mAP as a function of the rejection rate for DAP and DOV estimators computed for various \textit{N} values (2, 5, 10, 25 and 50). We chose these values of \textit{N} since we look for an order of magnitude of \textit{N} rather than a precise value. The idea is to know if we need a very large number of predictions to obtain a significant variance between the predictions, or if only a few predictions are sufficient. In addition, we did not go beyond 50 predictions because we want to keep a reasonable time and computational cost.

The results are given on Horae-test-300 for the page detection. The reject curves are constructed by first sorting the images according to their estimated confidence, the samples having a lower DAP value (or higher DOV value) than a pre-defined threshold are removed from the evaluation set and the mAP is computed over the remaining samples. For DAP, the threshold varies from 0 to 1 with a step of 0.05. For DOV, the values are not bounded, so the threshold varies from 10 down to 0 with a step of -1.

These plots show that using \textit{N}=10 predictions for the dropout estimation is enough, and that no improvement is observed with \textit{N}=25 or 50. Moreover, the computation cost is reduced with only 10 predictions. Based on this observation, we used \textit{N}=10 predictions with dropout to estimate the confidence scores in the rest of the experiments. 

\begin{figure}[t]
  \centering
  \includegraphics[width=\linewidth]{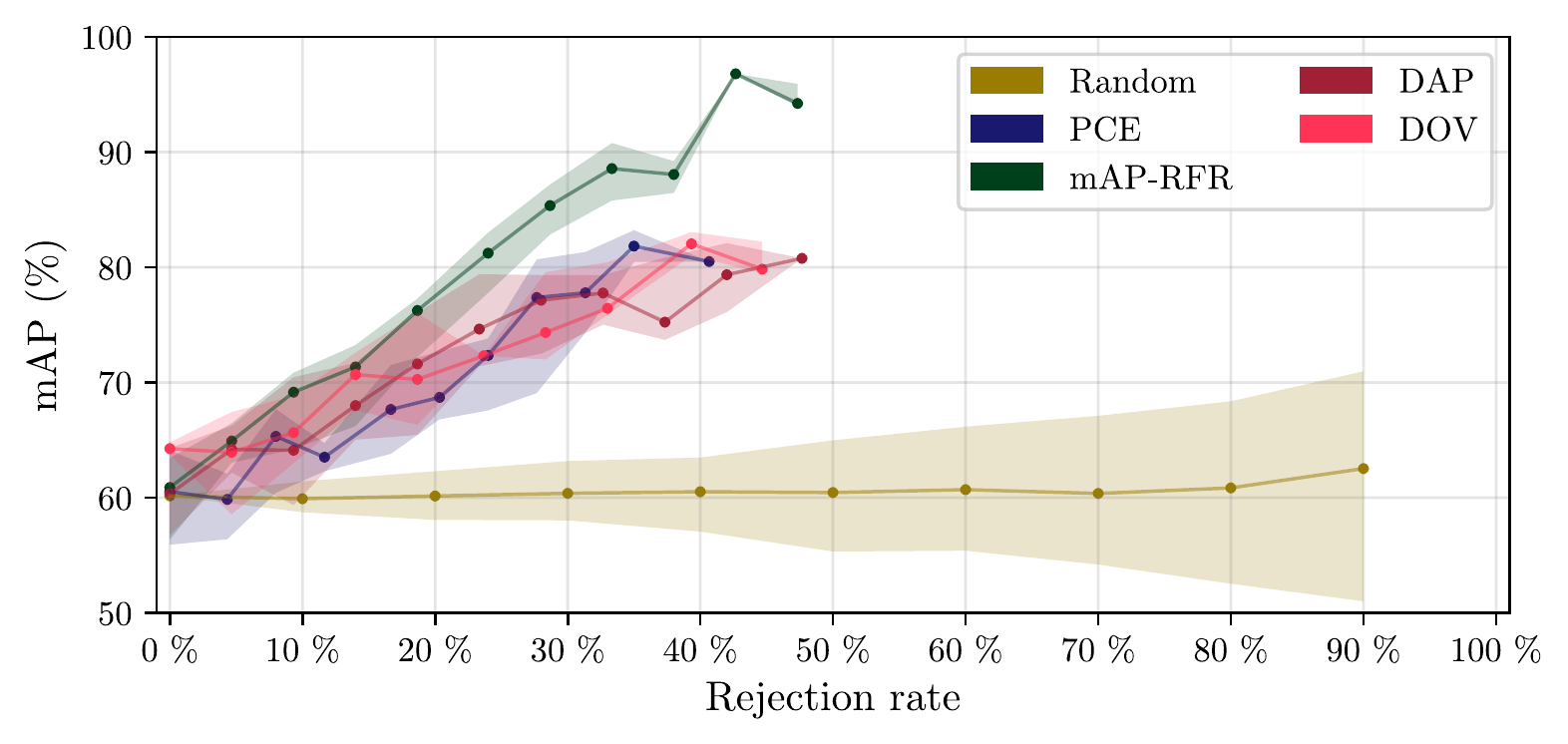}
  \caption{Reject curves showing the mAP score of the baseline page detection model on Horae-test-300 regarding the rejection rate.}
  \label{fig:rejection_curves_page_horae}
\end{figure}

\subsection{Confidence estimators performance in rejection}

This first experiment shows how the performance of the page detection model evolves when images with the lowest estimated confidence scores are removed from the test set. On reject curves, each point corresponds to a threshold, for which images with an estimated score below this threshold are removed from the evaluation. The curves do not reach 100\% because above a given threshold, only images with the same score remain so that they cannot be further removed without making the evaluation set empty. For the sake of clarity, we only show the evolution of mAP, the results of IoU following the same trend. \\

Figure \ref{fig:rejection_curves_page_horae} shows the evolution of the baseline model performance on Horae-test-300 for the page detection task regarding the rejection rate for different confidence estimators. We show the median curves as well as the confidence intervals (10th and 90th percentiles) obtained by computing 100 reject curves generated by 100 resamples with replacement from the original test set. The random curve shows the results obtained for 100 random samplings.

\begin{figure*}[t]
  \centering
  \includegraphics[width=0.59\linewidth]{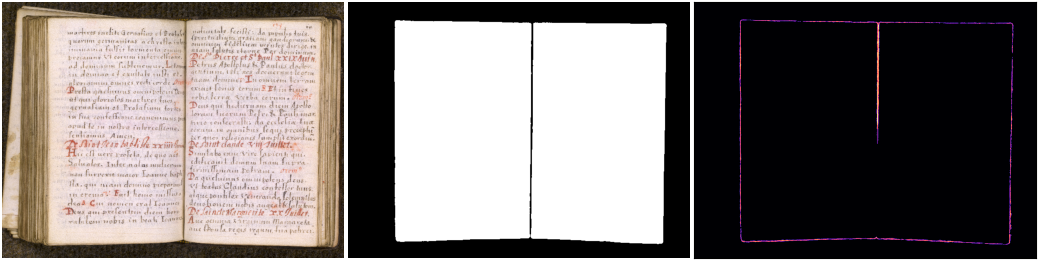}
  \hspace*{0.25cm}
  \includegraphics[width=0.28\linewidth]{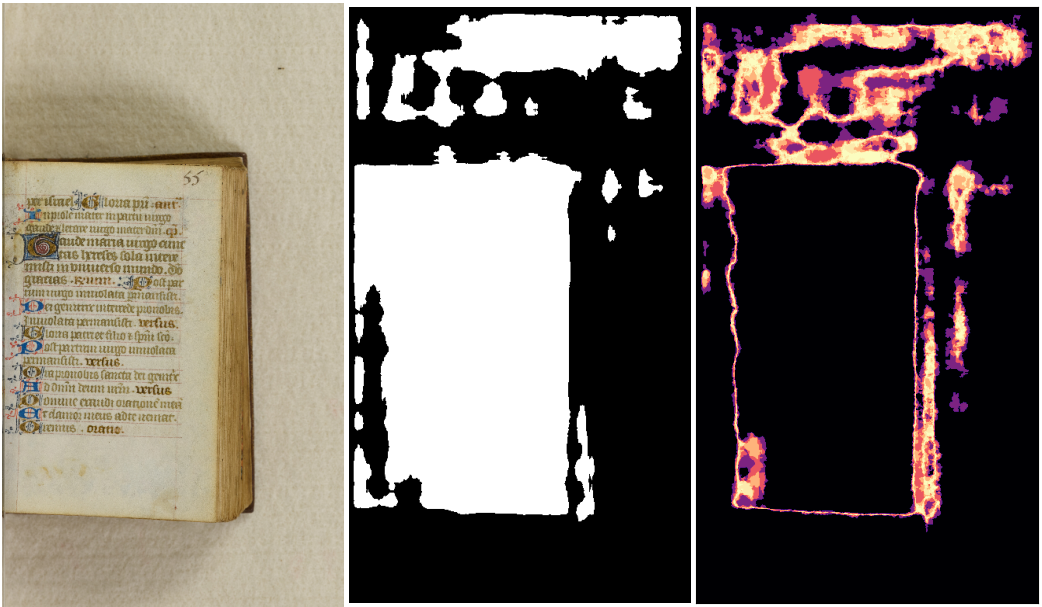}
  \caption{Two images from Horae (left) with their predictions (center) and the variance between \textit{N}=10 predictions with dropout (right). A high variance is represented in yellow, whereas areas with no variance are in black. The left image has confidence estimates of DOV=0.0, DAP=1.0 and mAP-RFR=1.0 and the right one DOV=17.36, DAP=0.0993 and mAP-RFR=0.5553.}
  \label{fig:variance_images}
\end{figure*}

The goal is to have a model with a high mAP and a low rejection rate. We can see that the dropout-based estimators do not hold up compared to the statistics-based regressor. Moreover, since mAP-RFR requires only one forward step in inference, this first result shows that using mAP-RFR instead of MC dropout is all the more interesting. The results of PCE being similar to DAP and DOV, it appears that the MC dropout estimators do not make better indicators than the detector posterior probabilities. This may be explained by the fact that no additional information than the trained neural network is provided to these three estimators.
This first experiment shows that our proposed mAP-RFR has a high ability to estimate the confidence of the predicted pages. It outperforms DAP and DOV which are themselves only slightly better than PCE. \\

On Figure \ref{fig:variance_images}, we show two predictions obtained by the baseline model for the page detection task. On the left we show a good prediction where the variance is very low except at the edges of the objects. The confidence estimates DOV=0.0, DAP=1.0 and mAP-RFR=1.0 reflect clearly the good quality of the detection of the left image while the confidence estimates of DOV=17.36, DAP=0.0993 and mAP-RFR=0.5553 of the right image also reflect the bad quality of the detection that contains a high variable number of small predicted objects around the main one. 

\begin{figure}[t]
  \centering
    \includegraphics[width=\linewidth]{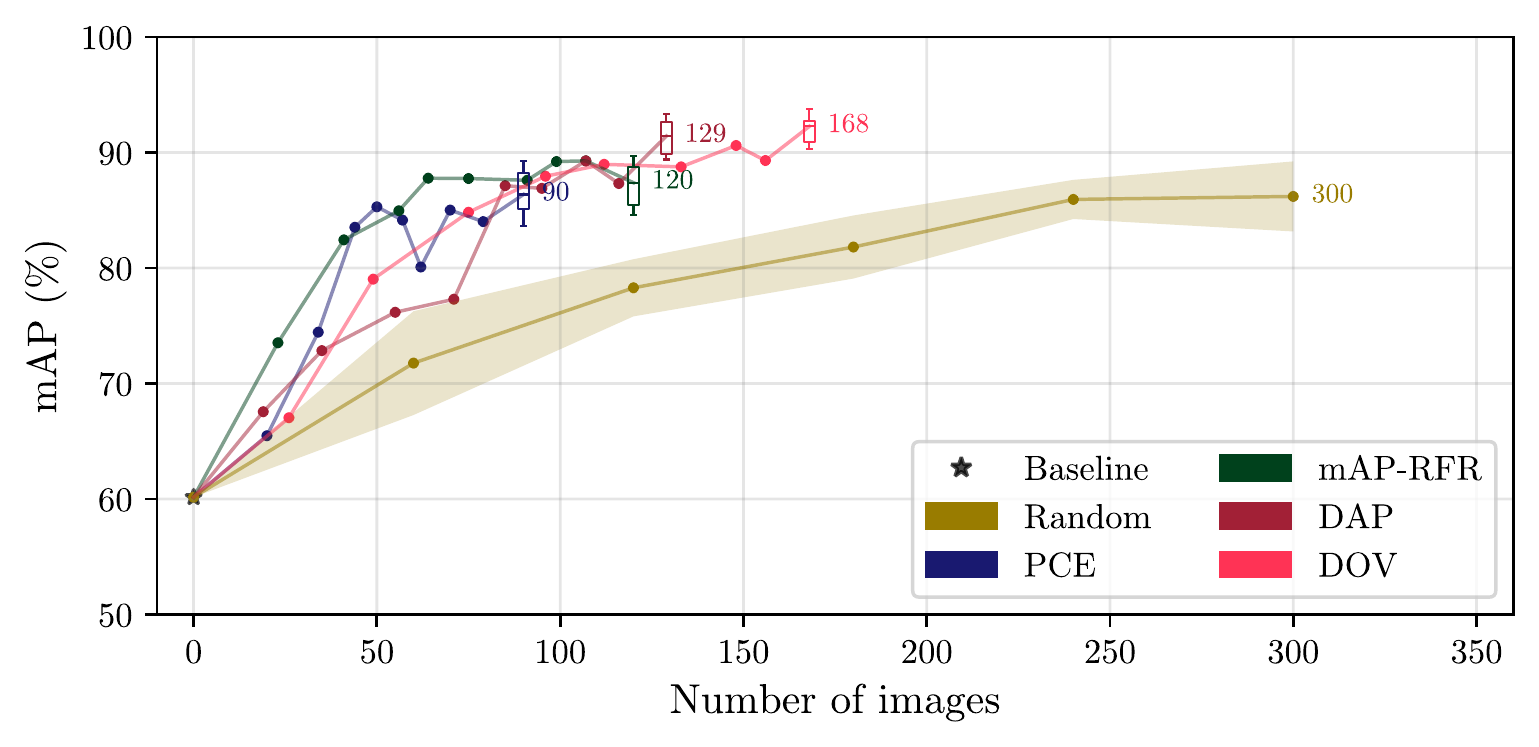}
    \caption{Evolution of the page detection evaluation (mAP) on Horae-test-300 during active learning iterations.}
    \label{fig:al_page}
\end{figure}

\subsection{Active learning}

Our main goal is to have a good object detector while requiring a reasonable amount of annotated samples. To achieve this, it is necessary to correctly choose which data to annotate.

In the experiments, we follow a standard active learning setup \cite{cohn1996}. First, a baseline Doc-UFCN model is trained and is then applied to unseen and non-annotated documents from a new dataset. Next, these images are ranked based on their estimated confidences, those with the lowest confidences are selected for manual annotation and used to train another model. Although many strategies for selecting the data to be annotated have been proposed to best improve the models \cite{settles2008}, we focus on selecting the images with the lowest confidence, other selection strategies are left for future works.

Each detection model is trained in the same configuration as the baseline models described in Section \ref{sec:page_segmentation_training}. During the active learning iterations, the models are fine-tuned and the weights are initialized with those of the last trained models. For the following experiments, we computed a confidence interval on the last models. For this, we used empirical bootstrapping \cite{wasserman2004} with 100 resamples with replacement. In addition, experiments with the random selection are repeated 5 times and the mean values and standard deviations are presented.

\begin{table}[t]
    \footnotesize
    \centering
    \caption{Results of the page detection models on Horae-test-300 after active learning}
    \label{tab:al_page}
    \begin{tabular}{lrrr}
        \hline
        \textsc{Estimator} & \textsc{IoU} & \textsc{mAP} & \textsc{\#images} \\
        \hline
        Baseline & 90.30 & 60.16 & 0 \\
        Random & 93.73 & 86.20 & 300 \\ \hline
        PCE & 93.45 & 86.37 & 90 \\
        mAP-RFR & 93.58 & 87.39 & 120 \\
        DAP & 94.04 & 91.43 & 129 \\
        DOV & 94.62 & 92.27 & 168 \\
        \hline
    \end{tabular} \\
    IoU and mAP are given in \%.
\end{table}

\subsubsection{Page detection}

Figure \ref{fig:al_page} and Table \ref{tab:al_page} show the results obtained for the page detection task. At each iteration, the last trained model is applied to the Horae corpus images and those with an estimated confidence score below a threshold are manually annotated and added to the training set. As with the rejection curves, these plots show that the estimators are able to detect bad predictions in order to train better performing models with only a small amount of annotated data. Indeed, they show that the estimators are better than a random selection since with twice less data, the models show relative increases of 6\% mAP (+5.23 percentage points) for DAP, 7\% (+6.07 points) for DOV and almost 1.5\% (+1.19 points) for mAP-RFR. From Figure \ref{fig:al_page}, we also observe that the curve corresponding to mAP-RFR is almost always above those of the other estimators, indicating better performing models with less annotated data.

These results confirm that mAP-RFR estimator outperforms MC dropout-based estimators since it shows higher mAP while requiring only one forward step during inference. One explanation for these results is that the DAP and DOV estimators are unsupervised: they have no prior knowledge about what a correct prediction is. On the contrary, mAP-RFR is trained with the real mAPs which are computed on the annotated data. 

\subsubsection{Text line detection}

\begin{figure}[t]
  \centering
    \includegraphics[width=\linewidth]{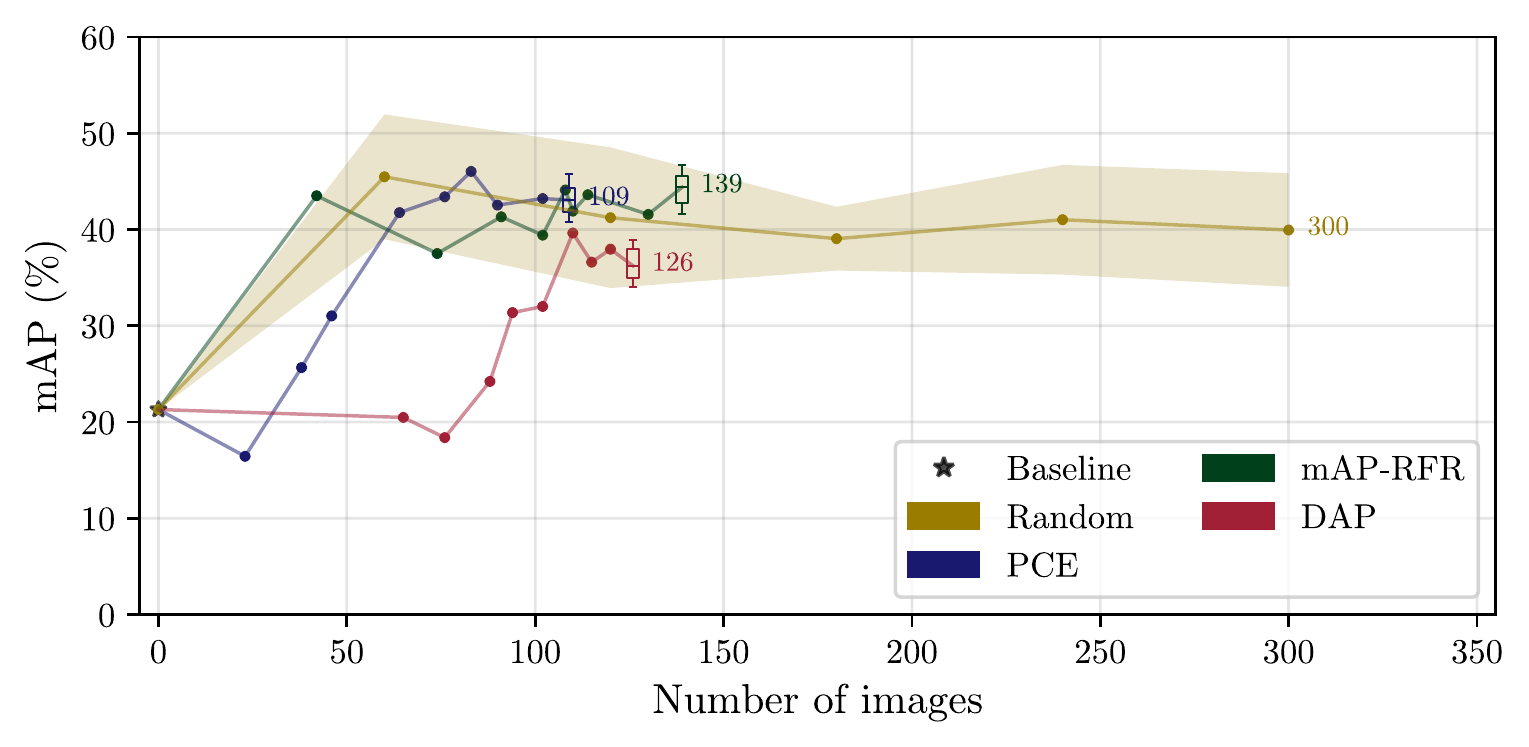}
    \caption{Evolution of the text line detection evaluation (mAP) on Hugin-Munin during active learning iterations.}
    \label{fig:al_line}
\end{figure}

Figure \ref{fig:al_line} and Table \ref{tab:al_line} show the results obtained when using mAP-RFR, DAP and PCE estimators for active learning. We do not show the results of DOV, since it shows very similar results to DAP. In addition, the WER is not shown here since it is strongly correlated to the CER. \\

According to Figure \ref{fig:al_line}, it appears that the random selection gives good results with only 50 images. However, these results strongly depend on the chosen data, which lead to highly variable performance from one selection to another. Therefore, in view of this large variability of the results, we believe that it is preferable to focus on a more robust and less random estimator that can obtain equally satisfactory results.

From Table \ref{tab:al_line}, mAP-RFR shows much better results than random data selection, with a relative increase of 11\% mAP (+4.4 percentage points) and a similar CER value with only half annotated images. It also shows a better IoU and mAP than the other estimators with a similar amount of annotated data. \\

In addition, despite much worse results in IoU and mAP, DAP shows better CER and WCFM values compared to mAP-RFR. Indeed, we show in \cite{boillet2022} that the relationship between these metrics is not linear. In addition, mAP-RFR was made to estimate the mAP of each prediction and thus maximize the mAP of the models. However, it has been shown that maximizing the mAP does not necessarily mean improving the input for the HTR \cite{boillet2022}. In fact, the Hugin-Munin dataset is annotated using quite complex polygons (including all ascenders and descenders). Therefore, improving the mAP consists in better predicting these ascenders and descenders which does not necessarily lead to the best input for the HTR. That is why mAP-RFR, which better predicts ascenders and descenders, shows worst text recognition results compared to DAP.

For this task, it would be worth selecting the images based on a confidence score related to text recognition. The detection model would directly adapt to improve the text recognition.

\section{Conclusion}

In this paper, we compared four approaches to estimate the confidence of object detection models. We have shown that these estimators can be used to train models reaching high performance for object detection in terms of IoU and mAP while requiring only a small manual annotation effort. When the optimized metrics are closely related to the final goal, such as for physical page detection, we have shown that mAP-RFR estimator leads to better detection performance than those based on MC dropout while having a reduced computational cost. However, it is supervised and needs to be trained, which is not the case for DAP, DOV and PCE. In the case of an adaptation to new data, it is therefore advantageous, as a first step, to use the dropout-based estimator DAP. If the results do not reach the expected performance, then it seems more valuable to use a trained estimator such as mAP-RFR. On the other hand, when the metrics are more loosely related to the final goal, such as for text line detection, dropout-based methods are more competitive.

\begin{table}[t]
    \footnotesize
    \centering
    \caption{Results of the line detection models on Hugin-Munin after active learning}
    \label{tab:al_line}
    \begin{tabular}{lrrrrr}
        \hline
        \textsc{Estimator} & \textsc{IoU} & \textsc{mAP} & \textsc{CER} & \textsc{WCFM} & \textsc{\#images} \\
        \hline
        Baseline & 48.21 & 21.30 & 24.37 & 59.35 & 0 \\
        Random & 61.96 & 39.95 & 22.20 & 63.26 & 300 \\ \hline
        PCE & 64.92 & 43.06 & 24.66 & 66.08 & 109 \\
        mAP-RFR & 64.36 & 44.39 & 22.50 & 65.73 & 139 \\
        DAP & 62.30 & 36.22 & 20.35 & 68.37 & 126 \\
        \hline
    \end{tabular} \\
    IoU, mAP and WCFM are given in \%.
\end{table}

In the future, we plan to estimate confidence at the object level directly in such a way that we no longer reject pages but objects. This would allow knowing exactly which objects are problematic and correcting them. In addition, we want to automatically create object description vectors through learned embeddings. Finally, we have shown that using goal-directed metrics allows evaluating the impact of the detection models on the final task. Therefore, we plan to set up an estimator that reflects the text recognition results.


\section*{Acknowledgments}

This work is part of the \textit{HOME History of Medieval Europe} research project, supported by the European JPI Cultural Heritage and Global Change (Grant agreements No. ANR-17-JPCH-0006 for France). It also benefited from the support of the Research Council of Norway through the 328598 IKTPLUSS HuginMunin project. Mélodie Boillet is partly funded by the CIFRE ANRT grant No. 2020/0390. Lastly, this work was granted access to the HPC resources of IDRIS under the allocation 2021-AD011011910 made by GENCI.


\bibliography{references}

\end{document}